\documentclass[11pt,letterpaper]{llncs}
\usepackage{amsmath}
\usepackage{amssymb}
\usepackage{graphicx}
\usepackage{marvosym}
\usepackage{url}
\usepackage{parskip}
\usepackage{fancyhdr}

\usepackage{geometry}
\newcommand{\keywords}[1]{\par\addvspace\baselineskip
\noindent\keywordname\enspace\ignorespaces#1}

\fancypagestyle{firstpage}{\fancyhf{}
\setcounter{page}{33}
\rfoot{\thepage}
}

\begin{document}

\title{
Subtractive mountain clustering algorithm applied to a chatbot to assist elderly people in medication intake
 }

\author{\large{Neuza Claro \and Paulo A. Salgado \and
T-P Azevedo Perdico\'ulis }}
\institute{\large{Escola de Ci\^encias e Tecnologia\\
Universidade de Tr\'as-os-Montes e Alto Douro,
Vila Real 5000--811, Portugal\\ Email: al68863@utad.eu, psal.utad@gmail.com, tazevedo@utad.pt}
}

\maketitle
\thispagestyle{firstpage}

\begin{abstract}
Errors in medication intake among elderly people are very common.   One of the main causes for this is their loss of  ability to retain information.  The high amount of medicine intake required by the advanced age is another limiting factor.
Thence, the design of an interactive  aid system, preferably using natural language, to help the older population with medication is in demand.  A chatbot based on a subtractive cluster algorithm, included in unsupervised learned models, is the chosen solution since the processing of natural languages is a necessary step in view to construct a chatbot able to answer questions that older people may pose upon themselves concerning a particular drug.
In this work, the subtractive mountain clustering  algorithm has been adapted to the  problem of natural languages processing. 
This algorithm version allows for the association of a set of words into clusters.  After finding the centre of every cluster — the most relevant word,  all the others are aggregated according to a defined metric adapted to the language processing realm.  All the relevant stored information is processed, as well as the questions, by the algorithm.   
The correct processing of the text enables the chatbot to produce answers that relate to the posed queries.
To validate the method, we use the package insert of a drug as the available information and formulate associated questions.
\end{abstract}

\section{Introduction}

Physicians often prefer the treatment of illness using medication.   One main reason for this is the non-invasiveness of this method of cure and another is the advance in science, namely in pharmacological engineering, that has given rise to new drugs and techniques with great effectiveness.

Older adults are more susceptible to the use of medication because they are also more prone to having chronic disorders such as high blood pressure, cardiac arrhythmia and diabetes. The use of multiple drugs to treat cumulative diseases --- multiple pharmacy --- and also the use of numerous medications to treat a single condition --- polypharmacy --- are very common situations among this age group.  A study carried out in the period 2010--2011 in the United States~\cite{usedrugs} showed that almost 90\% of elderly adults regularly take at least one prescription drug, approximately 80\% take at least two, and 36\% take at least five prescribed medications.

With the upsurge in medication, errors associated with their use have also raised, especially in older people~\cite{moreerrors}. The error associated with taking drugs is especially problematic for these generations because, besides being the ones with the most complex clinical conditions, they also have cognitive problems related to memory and assimilation of   information,  making it difficult to take the right medication at the right time and in the manner prescribed by their doctor. It is estimated that close to half of the older adults do not take their medication according to the doctor's  prescription~\cite{aginganddrugs}. 
Incorrect drug administration can happen by (i) taking medication at a different time of the day than it was prescribed;  (ii) taking a different dose from the prescribed one; and (iii) changing the medication in course on a particular occasion~\cite{medicationerror}.

These events may be caused by the similarity of the medicine box, the shape and colour of the pill, the similarity between the names of the medicines and the complexity and  length of the medical prescription~\cite{medicationerror}. One study confirmed that 25\% of the medication errors are associated with confusion with the name and 33\%  with confusion with the medication box and package insert~\cite{studymedicationerror}.
Medication intake errors can lead to loss of treatment effectiveness and increased risk of new complications may induce a new disease state, a new hospitalisation or even death.

Technology information in the health area has been growing~\cite{increasetechnology}. The use of systems to avoid medication errors at the hospital and home medication at home is vast. For example, Ahmed and coworkers~\cite{automaticdispenser} developed an automatic drug dispensing system. Mobile phone applications are also numerous~\cite{app}~\cite{app1}.

This work aims to develop an automatic conversational agent, a chatbot, capable of accompanying older people with taking medication. The person should be able to interact with this aid system in natural language.

The increase in the processing capacity of computers and smartphones, associated with the great development in artificial intelligence, has allowed the development of large\-scale software to facilitate the daily lives of humans. A bot, diminutive of a robot, is an automated hardware or software machine with the capacity to simulate human behaviour~\cite{bots}. Bots are powered by advances in Artificial Intelligence (AI) technologies. Inside a bot, an algorithm can produce a certain answer according to the input data. An example of a bot is the chatbot, a program capable of having an online conversation with a human being.

The evolution in machine learning algorithms, such as deep learning and deep reinforcement learning, has improved natural language processing  (NLP) performance. These advances have made intelligent conversational systems gain more and more popularity. Since then, chatbots have been applied in the most diverse areas: from commercial use~\cite{chatbotcommercial} to medicine~\cite{medicalchatbots}~\cite{10.3389/fmed.2019.00152}. 

Various NLP tasks are carried out to resolve the ambiguity in speech and language processing. Before machine learning techniques, all NLP tasks are carried out using various rules-based approaches.  In rule-based systems, rules were constructed manually by linguistic experts or grammarians for particular tasks. Machine learning and statistical techniques are everywhere in today's NLP. In literature, the implementation of various machine learning techniques for various  NLP  tasks has been investigated extensively. The machine learning systems start analysing the training data to build their knowledge and produce their own rules and classifiers. 
 
 Machine  learning techniques can be divided into three categories~\cite{machinelearning}~\cite{machinelearningNLP}:
  \begin{itemize}
      \item supervised learning whose models are trained using the labelled data set, where the model learns about each type of data, that includes models like hidden Markov model (HMM)~\cite{HMM} and support vector machines (SVM)~\cite{SVM}.
      \item semi-supervised learning that involves a small degree of supervision, and one example is bootstrapping~\cite{bootstrapping}.
      \item unsupervised learning whose model is not trained. The most common approach of the unsupervised category is clustering~\cite{clustering_1}~\cite{clustering_2}.
  \end{itemize}

Deep learning is a subfield of machine learning based on artificial neural networks which try to learn from the layered model of inputs. In the deep learning approach concept, learning of a current layer is dependent on the previous layer input. Deep learning algorithms can fall into both supervised and unsupervised categories~\cite{deeplearning}. The main applications of deep learning include pattern recognition and statistical classification. For example, to combat the increasing amount and reduce the threat of malicious programs, novel deep learning was developed~\cite{deeplearning_example}.

In  this work, we develop an aid system for medication intake --- a chatbot --- that   gives answers to queries related to the   medicine intake. Previously,  the system had to been informed about the user's prescription, that is, the  daily medicine intake routine and detailed information about every drug. We use the subtractive mountain clustering (SMC) algorithm to perform NLP tasks.

The main  objective of this work is  to apply the  SMC  algorithm within the chatbot to facilitate communication.  Namely to process natural language using the SMC algorithm.
 
The topics covered in this paper  relate mainly to chatbots and NLP. To conclude this section, we start by recalling some important facts in the history of chatbots in  Subsection~\ref{sec:history of chatbots} and in Subsection~\ref{sec:chatbots in healthcare} we review some existing chatbot solutions in healthcare. In Section~\ref{chatbotconfiguration}, the chatbot configuration is described. Then, in Section~\ref{NLP}, an overview of NLP and its implementation is discussed. In Subsection~\ref{subsubsec: preprocess}, we present the required steps for text pre-processing and,  in Subsection~\ref{subsec:response},  the SMC algorithm,  necessary to build the response, is described. Finally,  in Section~\ref{sec:results and discussion},  we   present and discuss the  results obtained for the case study.  We conclude with Section~\ref{sec: conclusion} where the main outcomes of the work are outlined and some guidelines for future work are stated.

\subsection{History of chatbots}
\label{sec:history of chatbots}

The first chatbot was created in 1966 by Joseph Weizenbaum.  It was called ELIZA~\cite{eliza}, and the objective was to pretend to be a psychologist. To do this, it used simple rules of conversation and rephrased most of what the users said to simulate a Rogerian therapist --- person-centreed therapy. In 1991, the Loebner Prize, an annual competition in artificial intelligence, was launched.  The contest awards the computer programs considered to be the most human-like. The competition takes the format of a standard Turing test, i.e., in each round, a human judge simultaneously holds textual conversations with a computer program and a human being via computer and, based upon the responses, the judge must decide which one is which~\cite{handsonchatbots}. In 2014, a chatbot named Eugene Goostman managed to fool 33\% of the judges, thereby beating the test.

Another example of a chatbot is using Natural Language Interface to Database (NLIDB) to access information in the databases instead of Structured Query Language (SQL).  An NLIDB system is proposed as a solution to the problem of accessing data in a simple
way: any user can access the information contained in the database and get the answer in natural language~\cite{NLIDB}. 
Nowadays, multiple virtual assistants already exist, being the most complex and more widely used. 
Siri (from Apple), Google Assistant (from Google) and Alexa (from Amazon).  At the moment, they are mainly used to call people, ask for directions or search the Internet for information~\cite{virtualassistants}.

\subsection{Chatbots in healthcare}
\label{sec:chatbots in healthcare}

These days, chatbots also have increased use in healthcare to treat disorders such as cancer~\cite{cancerchatbot} or induce behavioural changes such as quitting smoking~\cite{smokechatbot} and weight control~\cite{weighchatbot}. Chatbots are increasingly being adopted to facilitate access to information from the patient side and reduce the load on the clinician side.
In the field of medicine, there are already chatbots for the most varied purposes. Next, some illustrative examples are reviewed:

\begin{enumerate} 
 \item {\bf OneRemission} is a healthcare chatbot to help cancer survivors, fighters, and supporters to learn about cancer and post-cancer health care~\cite{oneremission}.
 \item {\bf Wysa} is an emotionally intelligent chatbot. Its purpose is to help the user to build mental resilience and promote mental well-being with a test-based interface~\cite{wysa}.
 \item {\bf Florence} is a chatbot related to medication intake that can remember taking medication, monitor certain biomedical parameters, and find information about diseases~\cite{helpchatbot}.
\end{enumerate}

After reviewing the chatbots already developed in healthcare, we did not find one that focuses on taking medication by older adults, like the chatbot presented here.

\section{Chatbot configuration}
\label{chatbotconfiguration}

The chatbot herein presented was designed to help older adults with their medication. The chatbot can provide information about the physician's prescription, the medicine's package insert, and also extra information, such as the colour of the box and the colour of the pill, to avoid confusion in taking medication.
Regarding the medical prescription, the chatbot can inform about the dose (how many pills per day) and when to take it (at which part of the day). Another set of information relates to the medicine's package insert, for example, indications, side effects, and what to do in case of forgetting to take it. Finally, the chatbot can also provide an image of the medicine box and the pill so that the patient can easily identify the medicine to be taken.  Once the medicine in question is identified, then the related information can be retrieved.

The set of questions and answers must be as similar as possible to a conversation between human beings. Chatbots are developed to connect with the users and feel that they are communicating with a human and not a bot.
In our chatbot, the possible answers are predefined and designed to emulate daily human communication.

Since a main step in the construction of the chatbot is   NLP, this topic will be discussed in Section~\ref{NLP}.

\section{Natural language processing}\label{NLP}
NLP is an area of computer science, more specifically, in artificial intelligence (AI),  concerned with giving computers the human ability to understand   text and spoken words. This processing generally involves translating natural language into data (numbers) that a computer can use and generate a certain answer. NLP is applied in several areas, such as, machine translation~\cite{machinetranslation}, text summarisation~\cite{textsummarization} and spam detection~\cite{spamdetection}. One of the NLP's uses is chatbots.

A chatbot system analyses a query posed by a person and generates an answer from   an organised collection of data stored and accessed electronically from a computer system.
  Usually, the answer is retrieved based on the basic keyword matching, and a selected response is then given as the output. When we talk about natural language, there are many ways to say the same information. So, when the chatbot is faced with several alternative sentences requesting the same information, it is necessary to use an algorithm that can import what is truly relevant. After selecting this information, the chatbot will be able to insert the phrase into a context  to produce the proper answer. Furthermore, the data needs to be pre-processed so that the chatbot can easily understand it.

\begin{figure}[t]
\centering
\includegraphics[width=7cm]{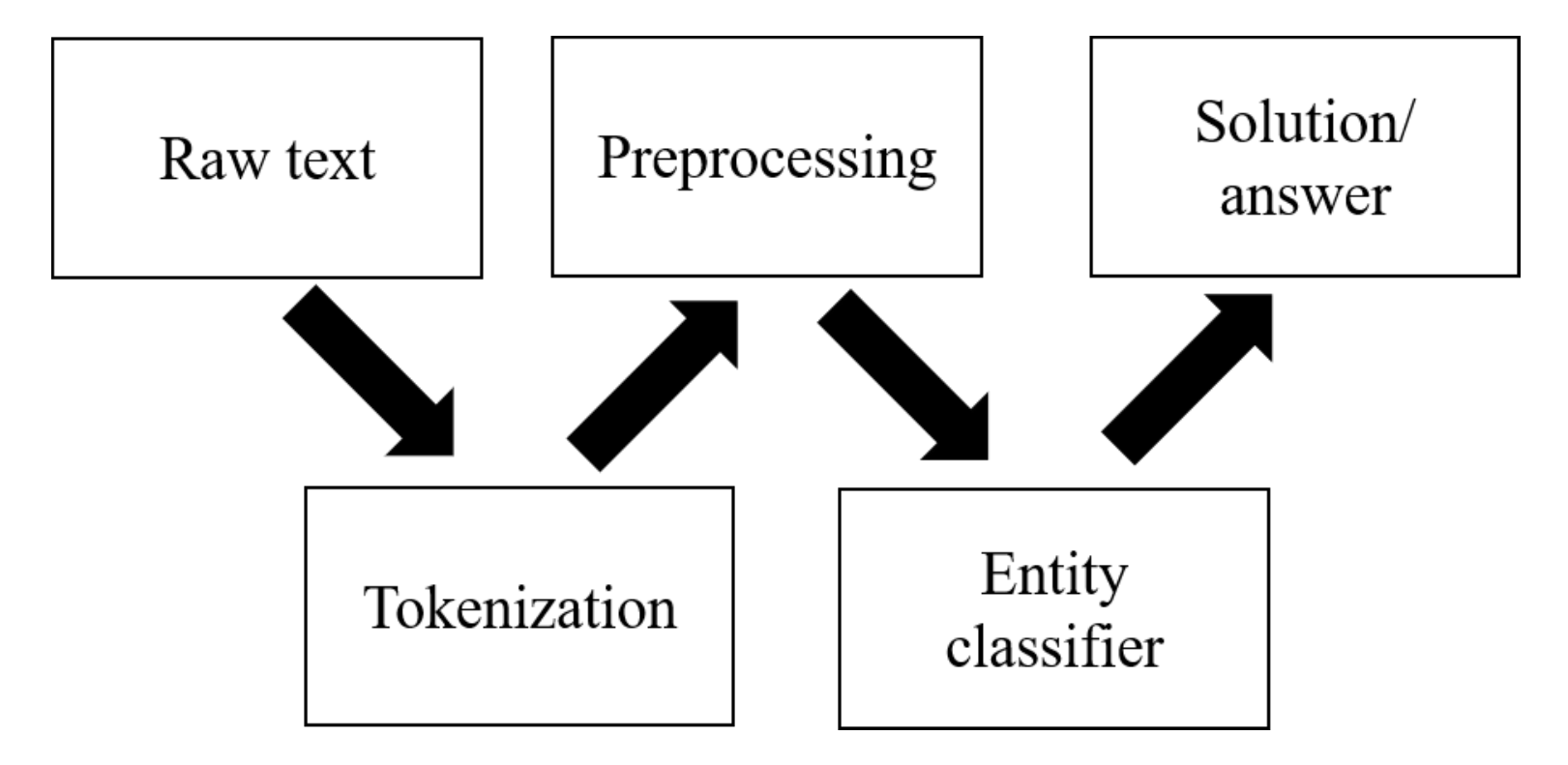}
\DeclareGraphicsExtensions.
\caption{Flowchart of the process needed from input to response in the chatbot.}
\label{fig:flowchart}
\end{figure}

In Figure~\ref{fig:flowchart}, one can see a flowchart that shows the process necessary for the chatbot to generate an answer fitting the user input text. In the following subsection, we will detail the pre-processing of text.  

We apply word processing to both (i) the drug's package insert,  which is used to define the algorithm that leads to the answers,  and (ii) the user's queries.

\subsection{Text pre-processing}
\label{subsubsec: preprocess}
Natural Language Understanding (NLU) converts natural language utterances into a structure that the computer can deal with. As a sentence cannot be processed directly to the model, it needs to perform some NLP to further operations. 

To achieve this, we implement an algorithm in MATLAB that uses a set of functions included in the {\it Text Analytics Toolbox.} Hence:

\begin{enumerate}
    \item Read the file with raw text.
    \item Segment the text into paragraphs --- documents.
    \item Remove empty documents.
    \item Segment the text into tokens ---
    tokenisation. This process consists in splitting the text into tokens, which are the basic units. 
    \item Remove unwanted words (stop words) or irrelevant punctuation from those tokens. Stop words are the words present in a sentence that does not make much difference if removed.  For example, the words and, off, for.
    \item Use normalising techniques, which consist in finding the root/stem of a word. For example, the words "ends" and "ending" are represented by the same stem - "end".
\end{enumerate}

The various steps above are not always an easy task to carry out. Non-standard words are often ambiguous. Some words can have different meanings depending on the context. Moreover, there are also some acronyms and abbreviations that can be misleading. For example, should an acronym be read as a word (IKEA) or using each letter in a sequence (IBM)? The abbreviation "Dr." has a full stop that can lead to the wrong separation of sentences.

The following step is word embedding, which consists in converting text to numbers. Converting words into numbers will make the algorithm easier to apply~\cite{preprocess}~\cite{preprocess2}~\cite{speechandlanguageprocessing}.

\subsection{Building the response}
\label{subsec:response}

Creating an algorithm that could relate the user's queries to the answers that contain the right information is a big deal for chatbots. 

Before selecting the correct answer is necessary to identify to which class/topic the query belongs. Here,  the SMC algorithm is used to define word sets (clusters)~\cite{mountaincluster}.

Based on the medication package's insert  is defined a proximity degree between the words related to medication intake. The calculus of the  proximity degree is based on the distance between any two words of the package insert.  Only relevant words are taken into account to calculate the distance. The word frequency and  size  are taken into account to evaluate relevance.  One possible criterion is, for instance, a word appearing more than twice and having more than two letters being classified as  relevant.

The calculus of the distance between two relevant words is based on their relative position in the text, where  $r$ and $s$ are the codes of the words that occupy positions $i$  and $j$, respectively. Hence, we distinguish three different cases: 
\begin{itemize}
    \item Two words are in the same sentence: the distance reflects the number of words that separate them. The end of  sentence recognition is done by punctuation marks. Then,  \(D(r,s) = \left|j-i\right|\).
    \item Two words are in the same document but in different sentences:   the distance reflects the number of sentences that separate them and the number of words. Let \(S_n\) be the number of separation sentences, then  their pair distance  is given by \(D(r,s) = \left|j-i\right|· S_n · a\), where \(a\) is an adjustable parameter according to the average number of words in each sentence.    
    \item Two words are in different documents: the distance only reflects the number of documents that separate them. Let \(P_n\) be the number of separation paragraphs, then their pair distance is given by \(D(r,s) = P_n · b\), where \(b\) is an adjustable parameter according to the number of words in sentences and sentences in the paragraph.    
\end{itemize}
This is indeed a distance, since is non-negative, symmetric and verify the triangular inequality.

As the words may appear many times, the minimum distance between them is considered. 

In addition, we also need a factor \(B(r,s)\) for each word pair that reflects the number of times each pair appears together in the same sentence and/or the same paragraph. 

In the  SMC  algorithm, we use the distance and \(B(r,s)\) to calculate the potential.  The algorithm is presented next.\\

\subsubsection{Subtractive mountain cluster algorithm}
\label{subsubsec: subtractive cluster}

Sometimes, it is necessary to reduce the size of the data set  to a set of representative points. For example, fuzzy logic algorithms are very complex and are not applicable in a large data set~\cite{inbook}.

The SMC approach, developed by Chiu~\cite{mountaincluster}, assumes that enormous data sets are partitioned into subsets called clusters, and each cluster is represented by one representative element called the cluster centre. Initially, all data set points are potential cluster centres and  each point potential is calculated by equation~(\ref{eq:potential}).  Hence

\begin{equation}
   P(I)=\sum_{j=1}^{N}e^{-\alpha{\left| x_i-x_j \right|^{2}}}, 
\label{eq:potential}
\end{equation}

for i = 1, …, N, (\(N\) is the number of points) and \( \alpha=4/(r_a)^{2}\) for constant \(r_a>0\). Equation~(\ref{eq:potential}) shows that a data point with many neighbouring data points will have a high potential value. Parameter \(r_a\) defines the data points influence. Data points outside this radius have little influence on potential.

 The SCM algorithm allows for the association of a set of words into clusters.  After finding the centre of every group — the most relevant term,  all the others are aggregated according to a defined metric adapted to the language processing realm. 

We adapted equation~(\ref{eq:potential}) to  NLP and define equation~(\ref{eq:potentialwithB}). So the greater the number of times a word pair appears, the greater the potential of that word pair.  Hence
\begin{equation}
   P(r,s)=B(r,s) \sum_{s=1}^{N}e^{-\alpha{{D(r,s)}^2}}, 
\label{eq:potentialwithB}
\end{equation}

where \(D\) is the symmetric distance matrix, \(N\) the number of words. We assume the potential of equals words pairs is approximately zero.

After the potential of every distance has been computed, as the matrix \(D\) is symmetric, we sum up all the potentials for each and every  word, and we obtained a vector with $N$ columns.  
Thus, we were able to obtain the potential for every word and choose the one with the greatest potential as the centre of the first cluster. Let \(I_1^*\) be the first word cluster centre index and \(P_1^*\) its potential value.

Equation~(\ref{eq:potential2}) is an adaptation of the potential equation for new cluster centre words after determining the first cluster centre word.  Hence

\begin{equation}
   P^{*}(r,s)= P(r,s) - P_1^* B(r,s) e^{-\beta{{D({I_1^*},s)}^2}},
\label{eq:potential2}
\end{equation}

 where \( \beta=4/(r_b)^{2}\) for constant \(r_b>0\). With Equation~\eqref{eq:potential2}, we subtract a portion of potential at each word pair according to the distance 
 from each word to   the word chosen as the centre of the first cluster. 
Words close to the first  cluster centre will have significantly reduced potential and are unlikely to be selected as the next cluster centre. Parameter \(r_b\) defines the   radius affected to the  potential reduction. 

Now, the word with more potential is selected as the second cluster centre. We then further reduce the potential of each word according to their distance to the second cluster centre. In general, each time we select the next centre of the next cluster,  we revise the value of the potential in the following manner:
\begin{equation}
   P_k^{*}(r,s)= P^{*}(r,s)- P_{k-1}^* B(r,s) e^{-\beta{{{D({I_k^*},s)}^2}}}.
\label{eq:potentialk}
\end{equation}

This iterative process ends when the word with the most potential \(P_k^*\) is less than \(\epsilon P_1^*\) where \(\epsilon\) is a small fraction. 

The minimum distance \(d_{min}\) between every two cluster centres also needs to be defined. If the following inequality is verified, the point is accepted as a possible centre of a cluster:
$$
    \frac{d_{min}}{r_{a}}+\frac{P_k^*}{P_1^*}\geq1 .
$$

Next,   the belonging degree of each word to the cluster  is calculated using equation~(\ref{eq:belonging}):

\begin{equation}
    U(I) = \frac{1}{\sum_{k=1}^c{\left(\frac{\parallel x_i-c_j\parallel}{\parallel x_i-c_k\parallel}\right)}^{\frac{2}{m-1}}},
\label{eq:belonging}
\end{equation}

where \(m\) is the hyper-parameter that controls how fuzzy the cluster is.

After applying the SMC algorithm, we can get groups of words (clusters) according to the distance they appear in the medicine package insert. It is necessary to pre-process the text of the questions, extracting the relevant information. Therefore, we can attribute them to a cluster and generate an appropriate answer.

\section{Results and discussion}
\label{sec:results and discussion}

The parameters used are available in Table~\ref{parameters}. We used the Xarelto drug package insert to validate the algorithm, which is used to prevent blood clots. 
\begin{table}[h]

\renewcommand{\arraystretch}{1.3}
\caption{Parameters}
\label{parameters}
\centering
\begin{tabular}{|c|c|}
\hline
\textit{\textbf{Parameters}} & \textbf{ Value } \\ \hline
\(a\)                        &        10       \\ \hline
\(b\)                        &        20      \\ \hline
\(r_a\)                      &        12       \\ \hline
\(r_b\)                      &        14       \\ \hline
\(\epsilon\)                 &        0.1   \\ \hline
\(m\)                        &        2      \\ \hline
\end{tabular}

\end{table}

We obtained 297 documents (paragraphs) and 838 tokens (include words and two important punctuation marks).

\begin{figure*}[t!]
 \centering
  \includegraphics[width=\textwidth]{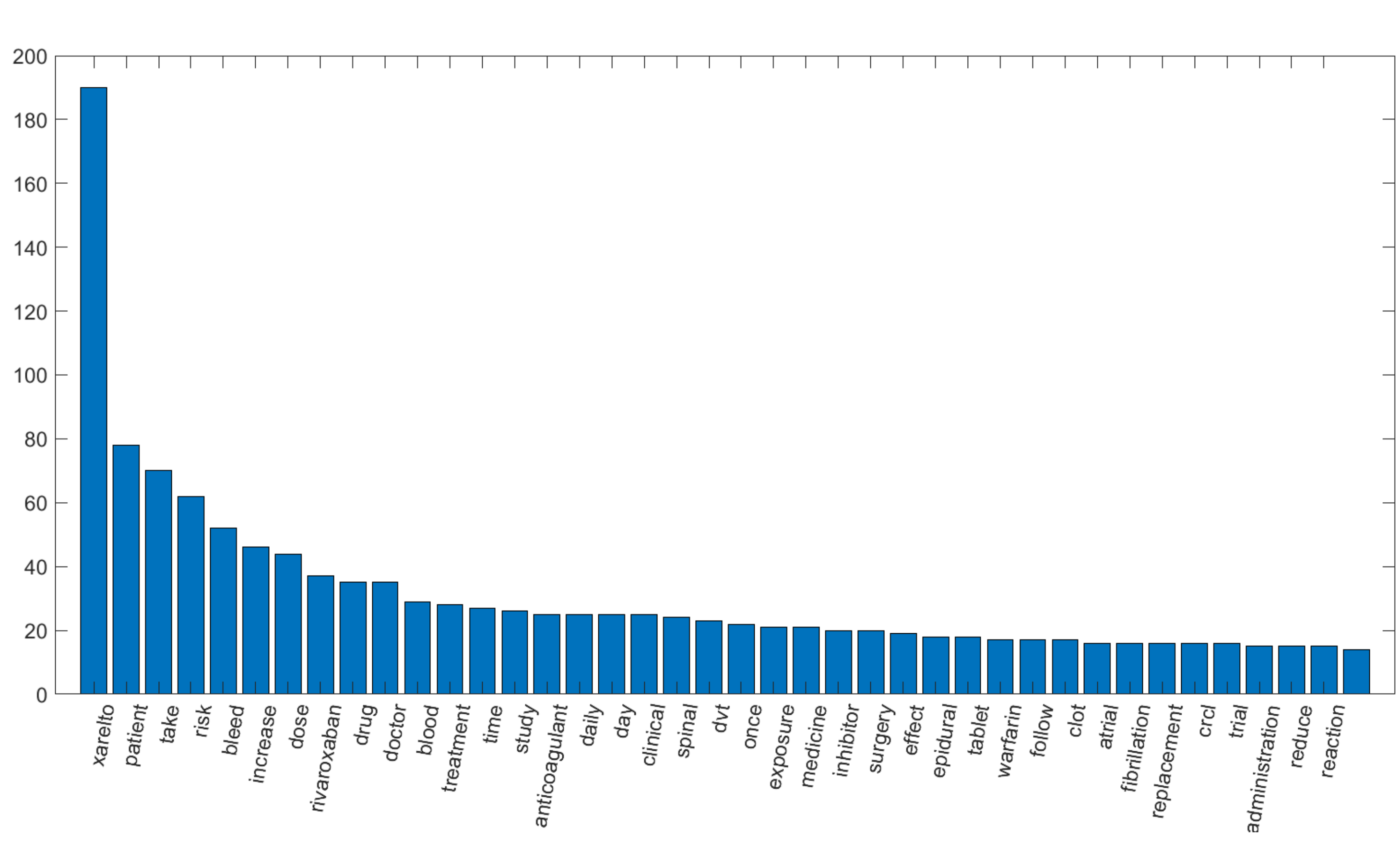}
  \caption{Number of times the first 40 most frequent words appear.}
  \label{fig:relevantwords}
\end{figure*}

Figure~\ref{fig:relevantwords} shows the number of times the first 40 most frequent words appear. The most frequent word is the name of the drug --- "Xarelto" and then "patient". After the first ten words, the frequency of each expression remains approximately the same.

After applying the word relevance metrics, we get 468 relevant terms representing 55.85\% of the total words.

\begin{figure}[h] 
\centering
\includegraphics[width=7cm]{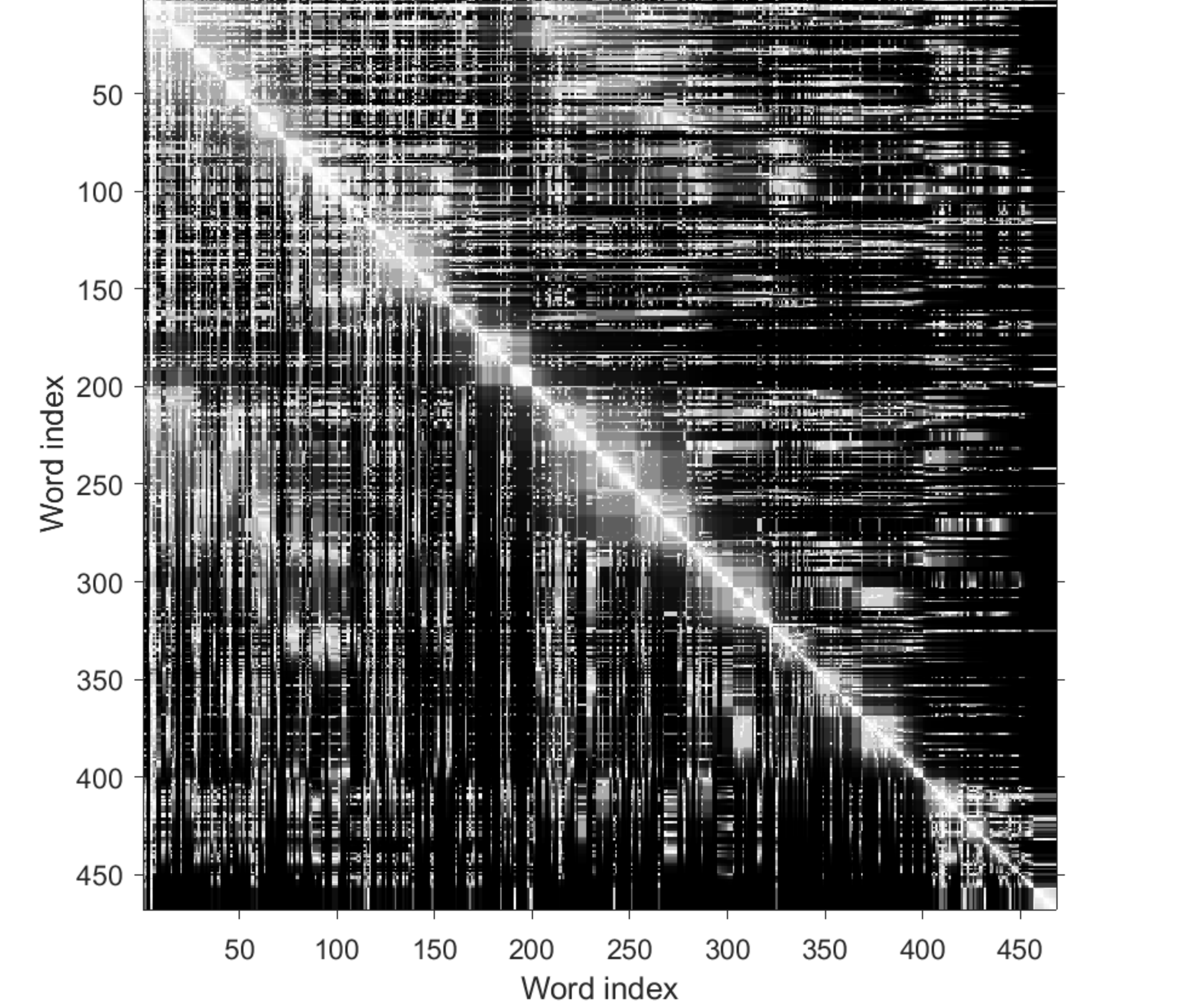}
\DeclareGraphicsExtensions.
\caption{Representative diagram of word distance. The white colour represents the closing words.}
\label{fig:distance}
\end{figure}

The distance between relevant words is represented in Figure~\ref{fig:distance}. On each axis, we have the indices of the words. The closer to white the colour, the smaller the distance between words. 

\begin{figure*}[h]
  \includegraphics[width=\textwidth,height=2.5cm]{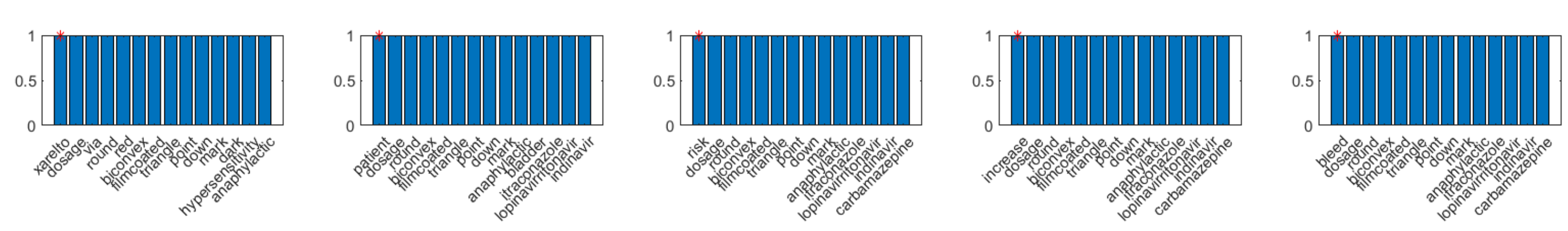}
  \caption{First five word clusters  in function of their normalising belonging degree. }
  \label{fig:clusters}
\end{figure*}

We get 20 clusters and the first five are represented in Figure~\ref{fig:clusters}. In the graphs, the degree of belonging is also represented. We can see that in each cluster there are at least 14 words with a degree of belonging equal to 1. 

\begin{figure}[t]
\centering
\includegraphics[width=\textwidth]{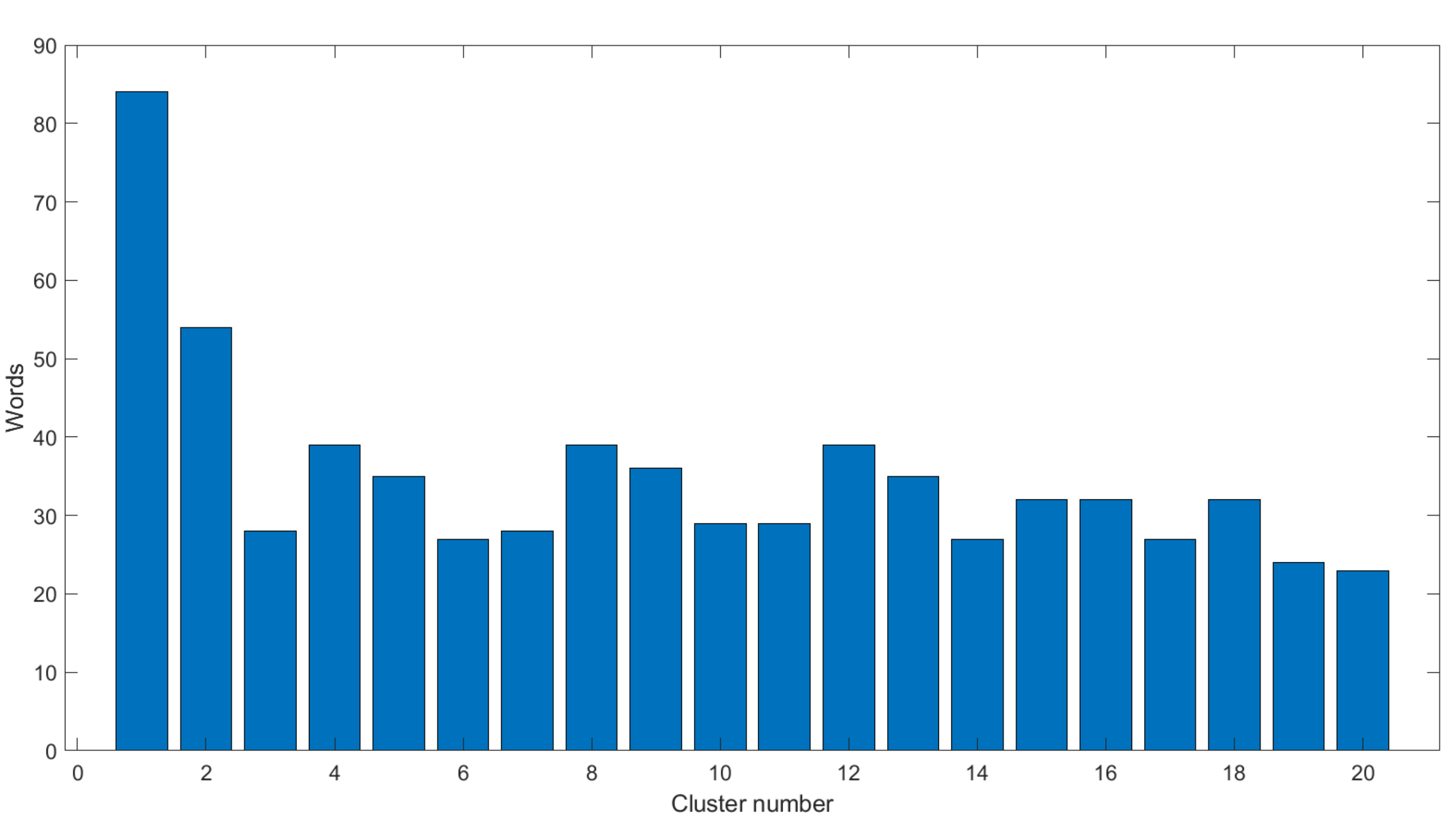}
\DeclareGraphicsExtensions.
\caption{The number of words belonging to each cluster with a degree of belonging greater than 0.5}
\label{fig:words}
\end{figure}

In addition, we calculate the degree of belonging of each word to the cluster and in Figure~\ref{fig:words} is the graph that shows the number of words belonging to each cluster with a degree of belonging greater than 0.5.

 We can get the most significant words, with question pre-processing, and associate them to a cluster or set of clusters. The answer is in accordance with the belonging relationship of words to clusters.
 
For example, one question with the stemming words "risk", "foetal" and "bleed" has multiple possible answers. We can sort answers by the relative relevance. Answers correspond to the paragraphs (documents) whose profile of belonging to the words in the question and the document are closest. The answer with 1.00 as relative relevance is "Xarelto increases the risk of bleeding and can cause serious or fatal bleeding. In deciding whether to prescribe Xarelto to patients at increased risk of bleeding the risk of thrombotic events should be weighed against the risk of bleeding."

\section{Conclusion and future work}
\label{sec: conclusion} 

Chatbots are present in different areas, including healthcare. In this field, there are chatbots for many purposes, from psychologist chatbots to medication reminders.

Chatbots have the potential to help elderly people in  taking medication  since they can emulate human conversation. However, after a thorough literature review, we conclude that they are still not widely used to solve and avoid medication errors.

A fundamental step to enable  communication between the users and  chatbots is the processing of natural languages.  
 The NLP area involves a large scientific community and contains a variety of associated algorithms.

We adapt the SMC algorithm to NLP. To the best of our knowledge, this algorithm has not been used before to resolve this type of problem.  This algorithm does not define {\it a priori} the number of clusters, leaving rather more freedom to the association.  Instead, through word clusters creation based on distance, we were able to define a degree of belonging among words and thus build adequate answers to the posed queries.

In this work, an aid system to help elderly people with medicine intake is proposed.   Namely, a chatbot that is able to interact with the user in natural language. To process natural languages, a fundamental step is to achieve a communication system-\ -user.  To do this,  we  apply the SCM algorithm to define the relationship between words. Furthermore, we implement  the proposed solution and applied it to a simple problem, obtaining results that we consider promising.

To complete this work, building a  user interface might be the next step.  The  refinement of  the algorithm is also in place. Moreover, the study of the accuracy of the SCM model should  be done as well as to investigate whether other algorithms can  solve the same problem. In addition,  a  study of the   adherence of people of more advanced age  to chatbots might be also an  interesting point of investigation.


\bibliographystyle{IEEEtran}
\bibliography{biomedica}

\vspace{2cm}

\section*{Authors}
\noindent {\bf Neuza Claro} is a Biomedicine 3rd student   at University of Trás-os-Montes  e Alto Douro.  She enjoys programming and being in   nature.  \\

\noindent {\bf Paulo A. Salgado}   received the B.S. degree and the M.S. degree in
electrical and computer science from the Faculty of Engineering,
University of Porto, Portugal, in 1989 and 1993, respectively, and the
Ph.D. degree in electrical and computer science from UTAD University,
Portugal, in 1999. He is currently an Associate Professor with the
Department of Engineering, School of Science and Engineering of UTAD, Portugal.
His research interests include artificial intelligence, control, and
robotics science.\\
 
\noindent {\bf T-P Azevedo Perdicoúlis} graduated in Mathematics (Computer Science) at the University of Coimbra in 1991, and pursued graduate studies at the University of Salford, Manchester (UK), where she obtained an MSc degree in Electronic Control Engineering in 1995 and a Ph.D. degree in Mathematics and Computer Science in 2000.  She works for the Universidade of Trás-os-Montes e Alto Douro,  Portugal,  since 1991 where she is currently an Associate Professor.   She is a founding member of the research institute ISR, University of Coimbra, and also of APCA, IFAC.  Her main research interests are in the fields of differential games, optimisation and simulation of gas networks  as well as identification theory.     Besides research she has got an intense pedagogic activity, having taught many different courses in applied mathematics to engineers. \\
     
\end{document}